\documentclass[letterpaper, 10 pt, conference]{ieeeconf}  

\IEEEoverridecommandlockouts                              

\usepackage[utf8]{inputenc}     
\usepackage[T1]{fontenc}
\usepackage{graphicx}
\usepackage{amsmath}
\usepackage{amssymb}
\usepackage{epstopdf}
\usepackage{flushend}
\usepackage[hidelinks]{hyperref}
\usepackage{soul}           
\usepackage{float}
\usepackage{multicol}
\usepackage[most]{tcolorbox}
\usepackage{cite}
\usepackage{multirow}
\usepackage[figuresright]{rotating}
\usepackage{placeins}
\usepackage{cancel}
\usepackage{url}
\usepackage{amsmath}
\usepackage{algorithm}
\usepackage[noend]{algpseudocode}
\usepackage{stfloats}
\usepackage[dvipsnames]{xcolor}
\usepackage{booktabs} 
\usepackage[table]{xcolor}
\usepackage{tikz}
\usepackage{bm}
\usepackage{physics}

\title{\LARGE \bf
Sliding into Balance: Benefits of a Sliding Payload for Load Distribution and Disturbance Rejection in Dual-UAV Transportation*
}

\author{Jorge Vicente-Martinez$^{1}$, Francisco-Jos\'e Mañas-\'Alvarez$^{2}$, Edgar Ramirez-Laboreo$^{1}$
\thanks{*This work is part of the projects \mbox{PID2024-159279OB-I00}, \mbox{PID2022-139187OB-I00} and \mbox{PID2024-156427OB-I00}, funded by MICIU/AEI/10.13039/501100011033 and ERDF/EU. It was funded also in part by the MICIU through the grant FPU24/01878 and in part by the Government of Arag\'on - EU, via grant T77{\_}26R.}
\thanks{$^{1}$Jorge Vicente-Martinez and Edgar Ramirez-Laboreo are with the Departamento de Informatica e Ingenieria de Sistemas (DIIS) and the Instituto de Investigacion en Ingenieria de Aragon (I3A), Universidad de Zaragoza, 50018 Zaragoza, Spain.
         {\tt\small j.vicente@unizar.es, ramirlab@unizar.es}}
\thanks{$^{2}$Francisco-Jos\'e Mañas-\'Alvarez is with the Department of Computer Science and Automatic Control of Universidad Nacional de Educación a Distancia (UNED), C/ Juan del Rosal 16, Madrid, 28040, Spain.
        {\tt\small fjmanas@dia.uned.es}}
}

\begin{document}

\maketitle
\thispagestyle{empty}
\pagestyle{empty}

\begin{abstract}
This paper analyzes the cooperative transportation of a payload suspended from two drones in which, instead of being attached to two independent ropes, the payload is allowed to slide along a single rope connecting both UAVs. First, a planar nondimensional dynamical model of the sliding motion with viscous friction is presented. Then, considering stationary drones, this configuration is compared with a fixed payload with the same total rope length, both at equilibrium and under disturbances. The analytical results show that the sliding payload equalizes the tension in both rope segments for any relative position of the drones, so that no rope loses tension at equilibrium; it doubles the region reachable by the payload and, except for low friction combined with a large drone separation, it enlarges the set of admissible disturbances. 
Finally, a dynamic simulation also shows that under realistic gusty wind profiles, the average tension imbalance between the drones is reduced with respect to the fixed-payload solution.
\end{abstract}


\section{Introduction}
In recent years, unmanned aerial vehicles (UAVs) have attracted growing interest for transporting payloads in logistics or emergency response, often to hard-to-reach locations~\cite{Villa2019JIRS, Estevez2024Drones}. Suspending the payload by cables is a mechanically simple and agile solution, although it poses significant planning and control challenges~\cite{Li2023RAL}. However, the capacity of a single vehicle is limited by its thrust, which has motivated the use of teams of drones that share the weight of the payload~\cite{Fink2011IJRR}.

In this regard, the literature on cooperative transportation is extensive. Early contributions addressed the static equilibrium and the inverse kinematics of the system formed by the drones, the cables, and the payload~\cite{Fink2011IJRR, Jiang2013TRO}, while later works have developed dynamical models and geometric controllers~\cite{Sreenath2013RSS, Lee2018TCST}, studied the role of internal forces in the stability of the equilibrium~\cite{Tognon2018RAL}, or interpreted the whole system as a reconfigurable cable-driven parallel robot~\cite{Masone2016IROS}. Distributed model predictive control~\cite{Wehbeh2020IROS}, control barrier functions for obstacle avoidance~\cite{Hegde2022LCSS}, and agile maneuvers~\cite{Sun2025SciRob} have also been proposed. In all of them, the payload is attached to each vehicle at a fixed point, either by cables or by rigid links. Thus, the load sharing is determined by the formation geometry: one drone may bear most of the weight, or a cable may lose tension~\cite{Jiang2013TRO, Tognon2018RAL}.

An alternative is not to attach the payload to the rope. Some works have exploited the interaction between cables held by drones and the object~\cite{DAntonio2021RAL, DAntonio2026TRO}: a cable suspended between two quadrotors can wrap around an object to manipulate it, or be interlaced with it to lift and transport it. However, they focus on grasping and maneuvering, and do not analyze how free sliding of the payload affects the load sharing and the response to disturbances, nor the role of friction at the contact.

\begin{figure}
\centering
  \includegraphics[]{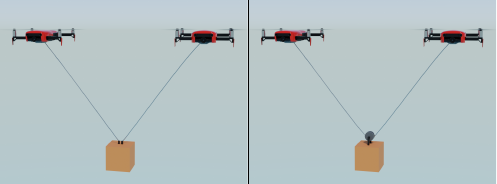} 
  \caption{Two-drone payload transportation: fixed-payload configuration with two ropes (left) and sliding-payload configuration with a single rope (right).}\label{fig:render_drones}
\end{figure}

In this paper we analyze, as a proof of concept, the effects of letting the payload slide along a single rope connecting both drones versus attaching it to two ropes with the same total length (Fig.~\ref{fig:render_drones}). 
A preprint published during the writing of this manuscript~\cite{DAntonio2026ARXIV} has already introduced this concept. 
However, our work explicitly considers friction and, additionally, presents results that clearly demonstrate the benefits of this solution.
To isolate the analysis from the controller, the drones are considered fixed points and the study is restricted to a vertical plane. The main contributions~are:
\begin{itemize}
\item \textbf{Model:} a nondimensional Euler--Lagrange model of the sliding motion with viscous friction, whose conclusions are independent of the system scale.
\item \textbf{Static analysis:} the sliding payload equalizes the tensions for any drone position, so that each bears half the weight and no rope loses tension. Hence, the formation can be chosen freely without overloading either drone. Besides, the region reachable by the payload is doubled.
\item \textbf{Disturbance analysis:} the admissible disturbances are characterized, and friction is shown to set a trade-off between tension balance and payload displacement, which guides the design of the contact point.
\end{itemize}
Overall, sliding provides a simple way, with no additional actuators, to decouple load sharing from formation geometry.

\section{System Description}

We consider two drones carrying a point payload of mass $m$ in the X-Z plane, suspended by ropes under two scenarios:
\begin{itemize}
    \item \textbf{Sliding payload.} A single rope of length $L$ connects the two drones (see Fig.~\ref{fig:esquema_identif}). The payload hangs from that rope and can slide along it affected by viscous friction.
    The time-dependent distance from the payload to drone 1 is $\ell(t)$, and the distance to drone 2 is $L-\ell (t)$. 
    \item \textbf{Fixed payload.} The payload is suspended from drone 1 by a rope of length $L/2$ and from drone 2 by a separate rope of the same length.   
\end{itemize}

The positions of the payload, drone~1, and drone~2 are defined by their coordinates in the plane, i.e., \mbox{$\bm {r}_{p}(t)=[x_{p}(t),z_{p}(t)]^\top$}, $\bm {r}_{d1}=[x_{d1},z_{d1}]^\top$, and $\bm {r}_{d2}=[x_{d2},z_{d2}]^\top$, respectively.

The objective of this study is to analyze the sliding payload's effect on static and dynamic behavior by evaluating rope tensions and drone reaction forces at equilibrium and under payload disturbances.
To isolate the suspended system from drone dynamics and controller design, in this preliminary work drones are assumed to be at fixed positions. Furthermore, ropes are inextensible and massless, and it is assumed that $x_{d2}>x_{d1}$ without loss of generality.

\section{Dynamical Model}
\label{sec: modelo}

\subsection{Sliding payload model with stationary drones}

\subsubsection{Payload kinematics}

\begin{figure}[t]
\centering
  \includegraphics[width=1\columnwidth]{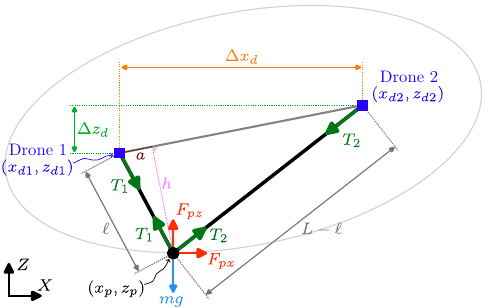} 
  \caption{Sliding-payload system in the $X$--$Z$ plane: a payload of mass $m$ slides along a rope of length $L$ held by two drones, dividing it into segments $\ell$ and $L-\ell$. The payload is located on the gray ellipse whose foci are situated at the drones. The payload is subject to its weight $m g$, the rope tensions $T_1$ and $T_2$, and a disturbance force $\bm F_p = [F_{px}, F_{pz}]^\top$. 
}
\label{fig:esquema_identif}
\end{figure}

Let $\Delta x_d=x_{d2}-x_{d1}$ and $\Delta z_d=z_{d2}-z_{d1}$ be the relative positions of drone 2 with respect to drone 1 along the X- and Z-axes, respectively, and let $d=\sqrt{\Delta x_d^2+\Delta z_d^2}$ be the Euclidean distance between them. Note that the rope length imposes the physical restriction \mbox{$d < L$}.
Based on the geometry of the triangles that are formed (see Fig.~\ref{fig:esquema_identif}), we can determine $a(\ell)$, the projection of $\ell$ onto the line connecting the drones, measured from drone 1, and $h(\ell)$, the distance from the payload to the line connecting the drones:
\begin{equation}
  a(\ell)=\frac{2L\ell-L^2+d^2}{2d}\,,
  \end{equation}
  \begin{equation}
      h(\ell)=\sqrt{\ell^2-a^2(\ell)}\,,
  \label{eq:ah}
\end{equation}
where the solution chosen ($h(\ell)\ge0$) corresponds to the payload hanging below the line connecting the drones. The payload position is then
\begin{equation} 
\bm r_p(\ell) = \begin{bmatrix}x_p(\ell)\\z_p(\ell)\end{bmatrix} = \begin{bmatrix}x_{d1}+ a(\ell)\, \tfrac{\Delta x_d}{d}\,+ h(\ell) \,  \tfrac{\Delta z_d}{d}\\z_{d1}+ a(\ell)\, \tfrac{\Delta z_d}{d}\,- h(\ell) \, \tfrac{\Delta x_d}{d}\end{bmatrix}\,.
  \label{eq:rc}
\end{equation}

\subsubsection{Euler--Lagrange dynamics}

The distance $\ell$ is chosen as generalized coordinate. The payload position is given by \eqref{eq:rc}, and its velocity is $  \dot{\bm{r}}_p= [\dot{x}_{p},\dot{z}_{p}]^\top =\bm J_p(\ell)\,\dot{ \ell}$ with $\bm J_p(\ell)=\partial\bm r_p/\partial \ell\in\mathbb R^{2\times1}$.
The Lagrangian is given by
\begin{equation}
 \mathcal{L} = \tfrac{1}{2} m \|\dot{\bm{r}}_p\|^2 - m g z_p,
\end{equation}
where $g$ is the gravitational acceleration.
The nonconservative generalized forces are the viscous friction during the motion of the payload along the rope, $Q_{\mathrm{fric}}(\dot\ell)=- b\,\dot\ell$, where $b$ is the viscous friction coefficient, and a disturbance force $\bm F_p=[F_{px},F_{pz}]^\top$ acting on the payload, which maps to the generalized coordinate as $Q_p=\bm J_p^\top(\ell)\bm F_p$.

Based on the Euler-Lagrange equation, 
\begin{equation}
    \frac{\mathrm{d}}{\mathrm{d}t}\left( \frac{\partial \mathcal{L}}{\partial \dot{\ell}} \right) - \frac{\partial \mathcal{L}}{\partial \ell} = Q_{\mathrm{fric}}(\dot\ell)+Q_p
    \label{eq: EL original}
\end{equation}
the equation for the payload's sliding motion along the rope can be obtained as
\begin{equation}
     M(\ell)\,\ddot\ell+N(\ell,\dot\ell) = Q_p\,,
  \label{eq:din_desliz}
\end{equation}
where $M(\ell) = \partial^2\mathcal L/\partial\dot{\ell}^2$, and
\begin{equation}
  N(\ell,\dot{\ell})=\frac{\partial}{\partial \ell}\!\left(\frac{\partial\mathcal L}{\partial\dot{ \ell}}\right)\dot{\ell}-\frac{\partial\mathcal L}{\partial \ell}-Q_{\mathrm{fric}}(\dot\ell)
\end{equation}
collects the Coriolis, centrifugal, gravity and friction terms.

The rope tension expressions can be obtained, e.g., by applying Newton's second law to the payload (see Fig.~\ref{fig:esquema_identif}):
\begin{equation}
  m\,\ddot{\bm r}_p=T_1\,\bm e_1(\ell)+T_2\,\bm e_2(\ell)- \,[0, m\,g]^\top+\bm F_p,
  \label{eq: newton}
\end{equation}
where 
  \begin{equation}
  \bm e_1(\ell)=\frac{\bm r_{d1}-\bm r_p}{\ell},\quad
  \bm e_2(\ell)=\frac{\bm r_{d2}-\bm r_p}{L-\ell},
  \label{eq:unitarios desliza}
\end{equation}
are the unit vectors from the payload to each drone, and $\ddot{\bm r}_p$ is the acceleration of the payload, given by
\begin{equation}
    \ddot{\bm r}_p = \frac{\partial\bm J_p(\ell)}{\partial\ell}\,\dot{\ell}^2\,+\,\bm J_p(\ell) \ddot{\ell}\,.
    \label{eq: ddrp}
\end{equation}

Then, the tensions are obtained as
\begin{equation}
  \begin{bmatrix}T_1\\T_2\end{bmatrix}=\bm{E}^{-1}(\ell)\left(m\,\ddot{\bm r}_p+  \;[0, m\;g]^\top-\bm F_p\right)\,,
  \label{eq:T}
\end{equation}
where $\bm E(\ell) = \begin{bmatrix}\bm e_1(\ell)&\bm e_2(\ell)\end{bmatrix}$. The model is valid for $h>0$ (at $h=0$, $\bm E(\ell)$ is singular) and $T_1,T_2>0$ (a negative value indicates a slack rope).
Drones have to counteract the tension in the ropes $\bm T_1 = T_1 \,\bm e_1(\ell)$, $\bm T_2 = T_2 \,\bm e_2(\ell)$. Therefore, these tensions can be used as an indication of the effort they must exert to maintain the position.

\subsection{Fixed payload model with stationary drones}

In the case of two independent ropes, as long as both ropes are under tension ($T_1,T_2>0$), a kinematic constraint is established that determines the position of the payload given the position of the drones. This system can be modeled as a particular case of the model presented in the previous section, where $\ell = L/2$ is constant and, thus, $\dot{\ell} = \ddot{\ell} = 0 \,. $

\subsection{Nondimensionalization}
The model above depends on $m$, $L$, $g$, $b$, $\bm r_{d1}$, and $\bm r_{d2}$, resulting in a high-dimensional parameter space that complicates systematic analysis.
To reduce the number of parameters, we nondimensionalize the system using the rope length $L$ as the characteristic length and $\sqrt{L/g}$ as the characteristic time. Spatial dimensions will be normalized as $\bar{(\cdot)} = (\cdot)/L$, and the rescaled time is $\tau = t\sqrt{g/L}$.
Hence, primes will denote differentiation with respect to $\tau$, yielding \mbox{$\mathrm{d}(\cdot)/\mathrm{d}t = \sqrt{g/L}\,(\cdot)'$}.

With $\ell=L\,\bar\ell$ and $\bm r_p=L\,\bar{\bm r}_p$, we have $\dot\ell=\sqrt{gL}\,\bar\ell'$ and $\dot{\bm r}_p=\sqrt{gL}\,\bar{\bm J}_p(\bar{\ell})\bar\ell'$, where the Jacobian is scale-invariant, 
$\bar{\bm J}_p(\bar{\ell})=\partial\bar{\bm r}_p/\partial \bar\ell = \bm J_p(L\,\bar\ell)$.
Let us define the scaled Lagrangian $\bar{\mathcal L}$ as 
\begin{equation}
  \bar{\mathcal L}=\tfrac12\|\bar{\bm J}_p(\bar\ell)\,\bar\ell'\|^2-\bar z_p \,.
\end{equation}
It can be easily shown that $\mathcal L=mgL\,\bar{\mathcal L}$.
The terms in \eqref{eq: EL original} can then be expressed in terms of $\bar{\mathcal L}$ as
\begin{equation}
  \frac{\mathrm d}{\mathrm dt}\!\left(\frac{\partial\mathcal L}{\partial\dot\ell}\right)
  = mg\,\frac{\mathrm d}{\mathrm d\tau}\!\left(\frac{\partial\bar{\mathcal L}}{\partial\bar\ell'}\right),
  \qquad
  \frac{\partial\mathcal L}{\partial\ell}=mg\,\frac{\partial\bar{\mathcal L}}{\partial\bar\ell},
\end{equation}
while the generalized forces can be expressed as $Q_p=\bar{\bm J}_p^\top(\bar\ell)\bm F_p$ and $Q_{\mathrm{fric}}(\bar\ell')=-b\sqrt{gL}\,\bar\ell'$. Dividing the whole equation by $mg$ yields
\begin{equation}
  \frac{\mathrm d}{\mathrm d\tau}\!\left(\frac{\partial\bar{\mathcal L}}{\partial\bar\ell'}\right)-\frac{\partial\bar{\mathcal L}}{\partial\bar\ell}
  =\bar{\bm J}_p^\top(\bar\ell)\bar{\bm F}_p-\bar b\,\bar\ell',
  \label{eq:EL_nondim}
\end{equation}
where
\begin{equation}
  \bar{\bm F}_p=\frac{\bm F_p}{mg},\qquad
  \bar b=\frac{b\sqrt{gL}}{mg}=\frac{b}{m} \sqrt{\frac{L}{g}}.
\end{equation}
Hence, $m$, $g$ and $L$ disappear from the model, and the system is fully characterized by the relative position of the drones $(\Delta\bar x_d,\Delta\bar z_d)$ with $\bar d<1$, the friction coefficient $\bar b$ and the disturbance $\bar{\bm F}_p$.

Expanding \eqref{eq:EL_nondim} in the same way as \eqref{eq:din_desliz}, we obtain
\begin{equation}
  \bar M(\bar\ell)\,\bar\ell''+\bar N(\bar\ell,\bar\ell')=\bar Q_p,
\end{equation}
where $\bar M=\partial^2\bar{\mathcal L}/\partial\bar\ell'^2$, $\bar Q_p=\bar{\bm J}_p^\top(\bar\ell)\bar{\bm F}_p$ and
\begin{equation}
  \bar N(\bar\ell,\bar\ell')=\frac{\partial}{\partial\bar\ell}\!\left(\frac{\partial\bar{\mathcal L}}{\partial\bar\ell'}\right)\bar\ell'-\frac{\partial\bar{\mathcal L}}{\partial\bar\ell}-\bar Q_{\mathrm{fric}}(\bar\ell').
\end{equation}
Given that $\bar{\mathcal L}$ does not depend on $m$, $g$, or $L$, the expressions obtained depend only on dimensionless variables and parameters.
Then, with $\bar T_i=T_i/(mg)$, these tensions are
\begin{equation}
    \begin{gathered}
\begin{bmatrix}\bar T_1\\ \bar T_2\end{bmatrix}=\bar{\bm E}^{-1}(\bar\ell)\left(\bar{\bm r}_p''+[0,1]^\top-\bar{\bm F}_p\right),\\
\bar{\bm r}_p''=\frac{\partial\bar{\bm J}_p(\bar\ell)}{\partial\bar\ell}\,\bar\ell'^2+\bar{\bm J}_p(\bar\ell)\,\bar\ell'',
\end{gathered}
\label{eq: tensiones adimensional}
\end{equation}
which is an equivalent formulation of \eqref{eq:T} with dimensionless parameters and disturbances, and where $\bar{\bm E}(\bar\ell)=[\bar{\bm e}_1\ \bar{\bm e}_2]$ has the same form as $\bm E$, with
$\bar{\bm e}_1=(\bar{\bm r}_{d1}-\bar{\bm r}_p)/\bar\ell$ and
$\bar{\bm e}_2=(\bar{\bm r}_{d2}-\bar{\bm r}_p)/(1-\bar\ell)$.

\section{Steady-State Analysis}

\begin{figure}
\centering
  \includegraphics[width=1\columnwidth]{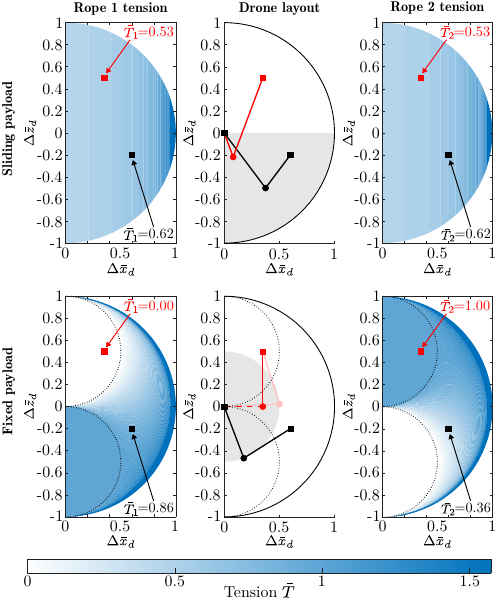} 
  \caption{Steady-state analysis of the sliding payload (top row) and fixed payload (bottom row) configurations. Lengths and tensions are normalized. The solid semicircle marks the maximum drone separation ($d=L$). The dotted circles delimit the region in which one of the ropes loses tension in that case. Two positions of drone~2 are highlighted: position~1, $(0.35,\,0.5)$, in red, and position~2, $(0.6,\,-0.2)$, in black. In the drone layout, the region reachable by the payload is in gray (at position~1 in the fixed-payload case, rope~1 becomes slack, dashed line, and the payload hangs vertically from drone~2, dark red, instead of occupying the position it would have with both ropes taut, light red).}\label{fig:analisis_estatico}
\end{figure}

In this section, we use the model from Section~\ref{sec: modelo} to conduct a steady-state analysis without external forces ($\bar{\bm F}_p=\bm 0$), demonstrating the advantages of the sliding payload configuration in terms of rope tension equilibrium and working zone area.
To do this, we compute the possible positions of the payload, the resulting normalized tension in each rope, and the forces the drones must apply to sustain them.
Drone~1 is placed at the origin, $\bar{\bm r}_{d1}=\bm 0$, and the variables are computed as a function of the position of drone~2 relative to drone~1, $(\Delta\bar x_d,\Delta\bar z_d)$ (Fig.~\ref{fig:analisis_estatico}).
We consider only $\bar{x}_{d2}> \bar{x}_{d1}$; the case $\bar{x}_{d2}<\bar{x}_{d1}$ follows by mirror symmetry about the vertical axis, and $\Delta\bar x_d=0$ is a degenerate limit. For the fixed payload, both ropes have length $\bar\ell=1/2$, so that the total rope length equals that of the sliding case; if one rope becomes slack, the payload hangs vertically from the other drone, which supports its whole weight ($\bar T=1$).

The maps show that with the sliding payload, the equilibrium is in the lowest point of the ellipse with foci at the drones. The payload always lies below drone~1, within the quarter disc of unit radius centered at drone~1, of normalized area $\bar A_p=\pi/4$. With the fixed payload, the payload lies at a distance $1/2$ from drone~1 when both ropes are taut. When a rope is slack, the payload hangs from one drone, which fills the half disc of radius $1/2$ ($\bar A_p=\pi/8$). The reachable region of the sliding payload is therefore twice as large and, in
addition, is fully available with both ropes taut.

The sliding payload guarantees equal tension across both rope segments, yielding $\bar T_1=\bar T_2=1/(2\cos\gamma)$, where $\gamma$ is the angle of each segment with the vertical. Consequently, tension remains strictly positive for any position of drone~2, preventing rope slackness. As shown in Fig.~\ref{fig:analisis_estatico}, the tension reaches its minimum ($\bar T\to1/2$) when the drones are close together and approaches infinity ($\bar T\to\infty$) as both $\bar d\to1$ and the ropes tend toward the horizontal.
In contrast, with the fixed payload, the tensions are in general unequal and one rope loses tension whenever drone~2 lies inside either of the two half discs of radius $1/2$ centered at
$(0,\pm 1/2)$ (dotted lines). 
With the sliding payload, the equal tensions lead to forces of equal magnitude on both drones, with the same vertical component $1/2$ and opposite horizontal components. For identical drones, this means a balanced load, so that neither battery would run out before the other. With a fixed payload, unequal forces can cause each drone to make a different effort, resulting in faster battery drain. Furthermore, in the slack zone, a single drone bears the entire weight of the payload ($\bar F_z=1$), while the other bears none, which significantly increases the force exerted by the first drone.

\section{Response to Disturbances}
In this section we analyze the rope tensions and the forces resulting from applying a disturbance force $\bar{\bm F}_p$ to the payload while the drones remain stationary. To this end, we first present the response to a step force in the X-axis of the fixed and sliding system, and then we analyze the range of disturbances in both the X and Z axes that the system can withstand without losing tension.

\subsection{Response to a Step Disturbance Along the X-axis}

To evaluate the dynamic behavior of both configurations, we first analyze a step force applied along the X-axis. A disturbance $\bar{\bm{F}}_p = [0.5, 0]^\top$ is applied at $\tau=0$, a value which ensures that neither configuration experiences rope slackness.
For comparison, both systems share the same initial condition and relative positions between the drones: $\bar\ell(0)=0.5$, $\Delta\bar{x}_d = 0.5$, $\Delta\bar{z}_d = 0$.

For the sliding configuration, the friction is expressed relative to the critical friction coefficient $\bar b_\text{crit}$, for which the linearized sliding dynamics around equilibrium is critically damped. Note that $\bar b_\text{crit}$ depends on $(\Delta\bar x_d, \Delta\bar z_d)$; the initial configuration yields $\bar b_\text{crit} = 10.53$.

\begin{figure}[t]
  \centering
  \includegraphics[width=1\columnwidth]{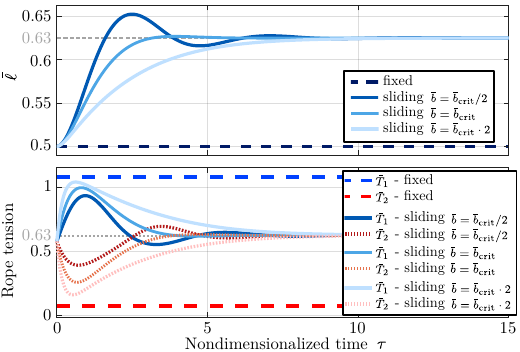}
  \caption{Response to a step disturbance $\bar{\bm{F}}_p = [0.5, 0]^\top$ at $\tau=0$ for $\Delta\bar{x}_d=0.5$ and $\Delta\bar{z}_d=0$, for both configurations under different friction coefficients $\bar{b}$. Top: time trajectory of $\bar{\ell}$. Bottom: evolution of the rope tensions $\bar{T}_1$ and $\bar{T}_2$.}
  \label{fig:escalon}
\end{figure}

Fig.~\ref{fig:escalon} (top) shows the payload displacement along the rope, given by $\bar{\ell}$.
The transient response is directly driven by the friction parameter $\bar{b}$: low friction leads to an underdamped behavior with overshoot and transient oscillations, whereas higher friction results in an overdamped response with longer settling time.

Fig.~\ref{fig:escalon} (bottom) presents the temporal evolution of the rope tensions $\bar{T}_1$ and $\bar{T}_2$. In the fixed payload configuration, the disturbance induces a permanent asymmetric tension, where rope 1 bears significantly higher tension than rope 2. In contrast, for the sliding-payload configuration, after the transient phase due to friction, tensions converge to equal steady-state values ($\bar{T}_1 = \bar{T}_2$).
In short, passive payload relocation along the ellipse guarantees equal steady-state tensions.

\subsection{Disturbances that Make the Ropes Lose Tension}

After analyzing the system's response to a step input, we analyze, for both systems, the step-type disturbance forces $\bar{\bm F}_p = \bar F_p[\cos\theta, \sin\theta]^\top$, where $\theta$ is the angle with the positive X-axis, that the system can withstand without losing rope tension.

\subsubsection{\textbf{Fixed payload}} 
Since the payload remains stationary ($\bar{\bm r}_p''=\bm 0$), \eqref{eq: newton} in normalized variables yields $\bar{\bm F}_p=[0,\,1]^\top-\bar T_1\,\bm e_1-\bar T_2\,\bm e_2$. Imposing $\bar T_1,\bar T_2>0$, the admissible disturbance region is bounded by
\begin{equation}
\begin{split}
    2(\bar x_{d2}-\bar x_p)+\bar{F}_p\sin{\left(\alpha_2-\theta\right)} >0, \\
    2(\bar x_{p}-\bar x_{d1})-\bar{ F}_p\sin{\left(\alpha_1-\theta\right)} >0,
\end{split}
\label{eq:cone_disturbances}
\end{equation}
where $\alpha_1$, $\alpha_2$ are the angles of $\bm e_1$, $\bm e_2$ from the positive X-axis. 
By solving for $\theta$, a convex cone of admissible forces with vertex at $(0, 1)$ bounded by $-\bm e_1$ and $-\bm e_2$ can be obtained (see Fig.~\ref{fig:juntas_zona_perturbaciones_SLIDING}).

\subsubsection{\textbf{Sliding payload}} 

\begin{figure}
\centering
  \includegraphics[width=1\columnwidth]{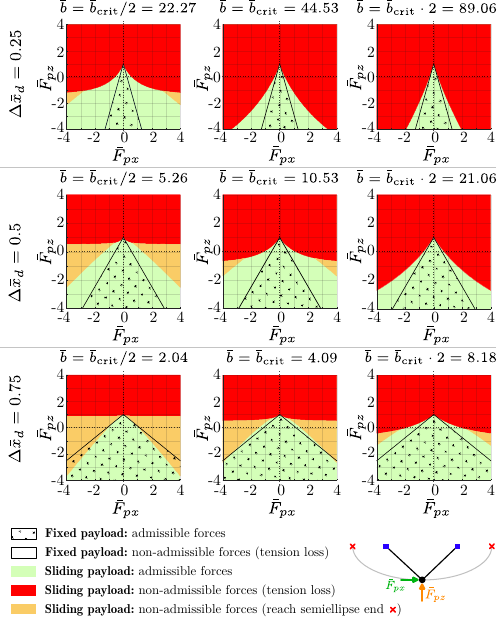} 
  \caption{Admissible disturbance regions for 3 different geometric configurations ($\Delta\bar z_d = 0$). Each is evaluated using its critical friction, twice that value, and half that value.}\label{fig:juntas_zona_perturbaciones_SLIDING}
\end{figure}

In the case of a sliding load, the applied disturbance force step causes a motion that induces a tension difference $\bar T_1-\bar T_2 = \bar b\,\bar\ell'$. A rope loses tension whenever $\bar b\,|\bar\ell'| > \bar T_1+\bar T_2$. Furthermore, in this configuration, the load may reach the end of the lower semiellipse (points marked with a red cross \textcolor{red}{$\boldsymbol{\times}$} in the diagram in Fig.~\ref{fig:juntas_zona_perturbaciones_SLIDING}). This scenario should be avoided, as the payload reaches its workspace limit, causing the rope to interfere with the drones.

To analyze this, we simulate the system's response to a step disturbance input for different geometric configurations. The drones are at the same height ($\Delta\bar z_d=0$) 
at separations $\Delta\bar x_d= 0.25, \ 0.5,$ and $0.75$, with critical frictions $\bar b_\text{crit} =44.53,\ 10.53$ and $4.09$, respectively; sliding payloads are then evaluated for $\bar b  = \bar b_\text{crit}/2,\ \bar b_\text{crit}$ and $2\,\bar b_\text{crit}$. For each of these configurations, $160{,}000$ combinations of step-like disturbances (a $400\times400$ grid over $\bar F_{px},\bar F_{pz}\in[-4,4]$) are applied at $\tau=0$ from the equilibrium and held until $\tau=50$. 

Fig.~\ref{fig:juntas_zona_perturbaciones_SLIDING} presents the results, which analyze—for each simulation—whether the ropes remain taut, whether any of them lose tension, or, in the case of the sliding configuration, whether the load reaches the end of the lower semiellipse. 
For the fixed configuration, the admissible region of disturbances is the cone given by \eqref{eq:cone_disturbances}, marked with a dotted area in the plots.
The plots show that low friction increases the risk of the payload sliding so far that it reaches the end of the lower semiellipse, while very high friction brings the system closer to the behavior of a fixed payload. Except for very low friction combined with large disturbances and a large separation between the drones (e.g., $\Delta\bar x_d=0.75$ and $\bar b=2.04$), the region of admissible disturbances is larger with the sliding payload. In that exception, the payload slides so much that it reaches the end of the lower semiellipse for disturbances that a fixed payload would withstand.

Since a constant acceleration of both drones is equivalent, in the drones' frame, to a disturbance on the payload, these maps can also be read as the accelerations that the pair of drones can apply without losing tension. For instance, the vertex $(0,1)$ corresponds to free fall.

\vfill
\section{Gusty Wind Simulation}

\begin{figure}[t]
  \centering
  \includegraphics[width=0.96\columnwidth]{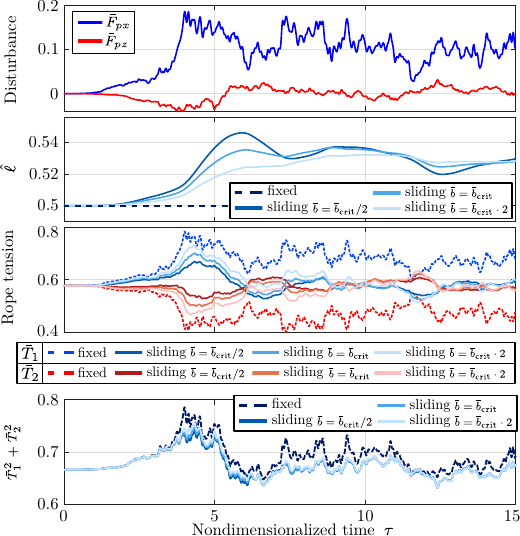}
  \caption{Response to a gusty wind. 
  First: wind force $\bar F_{px}$ (blue) and $\bar F_{pz}$ (red). Second: payload position $\bar\ell$ along the rope. Third: rope tensions $\bar T_1$ (blue) and $\bar T_2$ (red) for the fixed payload (dotted) and the sliding payload with $\bar b=\bar b_\text{crit}/2$, $\bar b_\text{crit}$ and $2\,\bar b_\text{crit}$ (solid, dark to light). Fourth: Power consumption proxy $T_1^2+\bar T_2^2$ for the four cases.}
  \label{fig:simula_viento1}
\end{figure}

This section evaluates the system's disturbance response in a scenario where stationary drones hover at $\Delta\bar{x}_d = 0.5$, $\Delta\bar{z}_d = 0$ with initial position $\bar{\ell}(0)=0.5$. The payload is subjected to wind drag ramping smoothly over $\tau\in[0,5]$ to mean values $\bar F_{px}\approx0.1$ and $\bar F_{pz}\approx 0$, with a superimposed Dryden turbulence component (Fig.~\ref{fig:simula_viento1}, first). The wind amplitude ensures the fixed configuration never loses tension, enabling a fair comparison.

For the fixed configuration, over $\tau\in[5,15]$ (i.e., after the initial ramp), the mean disturbance causes tension asymmetry ($\bar T_1 = 0.68$ vs.\ $\bar T_2 = 0.47$ on average), with turbulence peaks reaching $\bar T_1 = 0.78$ in rope 1 (Fig.~\ref{fig:simula_viento1}, third). Conversely, the sliding payload relocates along the rope (Fig.~\ref{fig:simula_viento1}, second), distributing tension more evenly. To quantify this imbalance, we define the root-mean-square tension difference:
\begin{equation*}
  \mathrm{RMSD}(\bar T_1,\bar T_2)=\sqrt{\left(\frac{1}{\tau_f-\tau_0}\int_{\tau_0}^{\tau_f}(\bar T_1-\bar T_2)^2\,\mathrm d\tau\right)},
\end{equation*}
which is zero for a perfectly balanced load. Compared to the fixed case, the sliding configuration reduces imbalance across all analyzed friction levels (see Table~\ref{tabla_resultados_viento}). At $\bar b=\bar b_\text{crit}$, $\mathrm{RMSD}(\bar T_1,\bar T_2)$ drops by a factor of $3.26$ and peak tension by $10.46\%$. Since $\bar{T}_1-\bar{T}_2=\bar b\,{\bar\ell}'$, tension imbalance grows with friction, whereas lower friction yields larger payload displacement and oscillations. Thus, $\bar b$ trades off tension balance against payload motion.

Since each drone must counteract the tension in its rope through control inputs, we also evaluate the mean force exerted by each drone on the rope computing the average of the tensions in each rope, ($\langle\bar{T}_1\rangle$, $\langle\bar{T}_2\rangle$) and the sum of squared tensions, $\bar T_1^2+\bar T_2^2$, as a proxy of the power consumption (Fig.~\ref{fig:simula_viento1}, fourth).
It also captures how the load is shared, since $\bar T_1^2+\bar T_2^2=\tfrac12\left[(\bar T_1+\bar T_2)^2+(\bar T_1-\bar T_2)^2\right],$
so for a given total tension it is minimal when the load is evenly distributed between the two ropes.
With the fixed payload, the mean tension of drone 1 is $1.37$ times that of drone 2, whereas with the sliding payload this ratio drops to \mbox{$1.018$--$1.071$}.
In the graph of the power consumption indicator, we can see that, in the case of a fixed rope, power consumption is higher most of the time than that of a sliding rope.
These results reveal two practical advantages. On the one hand, the maximum tension supported by the ropes is lower and evenly distributed, delaying fatigue in any single rope. And, on the other hand, the control effort is balanced between the drones preventing one drone from depleting its battery first.

\renewcommand{\arraystretch}{1.4}
\begin{table}[h]
\caption{Nondimensionalized tension metrics.\\Simulation under gusty wind ($\tau\in[0,15]$). }
\label{tabla_resultados_viento}
\begin{center}
\begin{tabular}{c|c|c|c|c|}
\cline{2-5}
                              & \multirow{2}{*}{Fixed} & Sliding & Sliding & Sliding \\
\multicolumn{1}{l|}{}         &                        & $\bar{b}_\text{crit}/2$     & $\bar{b}_\text{crit}$       & $2\,\bar{b}_\text{crit}$     \\ \hline
\multicolumn{1}{|c|}{$\mathrm{RMSD}(\bar T_1,\bar T_2)$}     & $0.203$                  & $0.0462$  & $0.0623$  & $0.0833$  \\ \hline
\multicolumn{1}{|c|}{max($\bar{T}_1,\bar{T}_2$)}     & $0.784$                  & $0.670$   & $0.702$   & $0.729$   \\ \hline
\multicolumn{1}{|c|}{$\langle\bar{T}_1\rangle$} & $0.667$                  & $0.585$   & $0.590$   & $0.600$   \\ \hline
\multicolumn{1}{|c|}{$\langle\bar{T}_2\rangle$} & $0.489$                  & $0.575$   & $0.570$   & $0.560$   \\ \hline
\end{tabular}
\end{center}
\end{table}
\renewcommand{\arraystretch}{1}

\newpage
\section{Conclusions}

We presented a proof of concept for planar payload transport using two drones connected by a single sliding rope rather than two separate fixed ropes.
Stationary-drone analysis shows three key advantages of the sliding configuration over a fixed payload of equal rope length.
First, passive payload relocation equalizes tension in both rope segments, balancing drone forces and—for identical drones—battery consumption. This advantage is particularly important when drones are not at the same altitude—for example, to avoid an obstacle.
In a realistic wind simulation with critical friction, the force ratio between drones can drop from $1.37$ to $1.03$ while total force remains practically unchanged.
Second, the sliding setup withstands a larger set of constant disturbances without tension loss, except under very low friction with large drone separation, where the payload reaches the end of the lower semiellipse.
Third, when the relative position of the drones changes (e.g., when only one of them moves), the region reachable by the payload is twice as large with the sliding configuration, and both ropes remain taut for any drone position, whereas with the fixed payload one rope loses tension in part of the space.

Drawbacks of the sliding configuration include payload shifting along the cable, meaning its location is no longer purely geometric and requires controllers to use a more complex model incorporating friction. Additionally, this analysis simplified the system by neglecting dry friction.

Future work will extend this analysis to 3D, incorporate drone dynamics, control, and trajectory planning, study active friction regulation (e.g., via a payload brake) to dynamically select the optimal friction level, and experimentally validate the results.

\addtolength{\textheight}{-12cm}   

\section*{Acknowledgement}
Gemini was used for language improvement; the authors reviewed all content and take full responsibility for the text.

\bibliographystyle{IEEEtran}


\end{document}